\documentclass[runningheads]{llncs}
\usepackage{graphicx}
\usepackage{amsmath}
\usepackage{amssymb}
\usepackage{xcolor}
\usepackage{algorithm}
\usepackage{algpseudocode}

\begin{document}
\title{Unsupervised Domain Adaptation with Progressive Domain Augmentation}
%
%
\author{
Kevin Hua \and
Yuhong Guo
}
\authorrunning{K. Hua and Y. Guo}
%
\institute{School of Computer Science, Carleton University, Ottawa, Canada\\
\email{kevinhua@cmail.carleton.ca, yuhong.guo@carleton.ca}}
\maketitle              
\begin{abstract}
Domain adaptation aims to exploit a label-rich source domain for learning classifiers 
in a different label-scarce target domain.  
It is particularly challenging when there are significant divergences between the two domains. 
In the paper, we propose a novel unsupervised domain adaptation method based on progressive domain augmentation.
The proposed method
generates virtual intermediate domains via domain interpolation, 
progressively augments the source domain and bridges the source-target domain divergence
by conducting multiple subspace alignment on the Grassmann manifold. 
We conduct experiments on multiple domain adaptation tasks and the results
shows the proposed method achieves the state-of-the-art performance. 

\end{abstract}
\section{Introduction}
Domain adaptation has seen a sharp rise in techniques and research over the last few years \cite{Li2019Cycle-consistentNetworks,YuTransferNetwork}, due in part to the longstanding issue of lacking sufficient amount of labeled data \cite{WangDeepSurvey}. 
In this age of technology, vast amounts of unlabeled, raw data is constantly being created in almost every domain; unfortunately, annotating this data is an expensive undertaking \cite{Beijbom2012DomainApplications}. 
Domain adaptation addresses the lack of data labels by leveraging labeled data from some related source domain 
to aid in learning a classifier for a label-scarce target domain. 
There are three primary types of domain adaptation: supervised (small amount of labeled target data), semi-supervised (limited labeled target data with redundant unlabeled data), and unsupervised (no labeled target data) \cite{WangDeepSurvey}; this paper focuses on the unsupervised domain adaptation setting.

Traditionally, models are trained with the understanding that the training instances and testing instances are drawn from the same probability distribution \cite{Tommasi2012LearningKnowledge}. However, this is not the case with domain adaptation tasks; the assumption instead is that the two domains have different probability distributions. This mismatch of training (\textit{source}) and test (\textit{target}) data distributions, $P_{\text{source}}(x) \neq P_{\text{target}}(x)$, is also referred to as the domain divergence or covariate shift \cite{Kumar2018Co-regularizedAdaptation,Shimodaira2000ImprovingFunction}; the goal of domain adaptation then becomes to learn some model that performs well on the target domain by transferring relevant knowledge learned from the source domain \cite{Zhang2013DomainShift}. The problem of domain shift is particularly common in visual recognition or classification tasks, as different cameras are used, different lighting conditions, and even different angles \cite{Baktashmotlagh2013UnsupervisedProjection}.
Many approaches have been developed to carry out domain adaptation; a large portion of them choose the route of exploring domain invariant structures or representations to induce aligned distribution across the source and target domains \cite{Zhao2019OnAdaptation,Ganin2014UnsupervisedBackpropagation,Hoffman2013EfficientRepresentations,Gu2019ImprovingInformation}. This particular category of approach bears many angles; domain invariant features can be learned via the minimization of divergence between the distributions of the two domains \cite{Long2015LearningNetworks,Sun2015ReturnAdaptation,Kang2019ContrastiveAdaptation,Damodaran2018DeepJDOT:Adaptation,Saito2018MaximumAdaptation,Zellinger2017CentralLearning}, via adversarial training \cite{Liu2016CoupledNetworksb,Ganin2017Domain-adversarialNetworksc,Hoffman2017CyCADA:Adaptationb,Ganin2014UnsupervisedBackpropagation}, or via an auxiliary reconstruction task \cite{Ghifary2016DeepAdaptationb,Fernando2014SubspaceAdaptation,Thopalli2019MultipleAdaptationb,Zhu2017UnpairedNetworksc}.
Although such methods that directly align the source and target domains
have demonstrated good adaptation performance, they can encounter great challenge 
and cause information loss \cite{Morerio2020GenerativeAdaptation}
when the cross-domain divergence is large.

In this paper, we propose a novel Progressive Domain Augmentation (PrDA) method
for unsupervised domain adaptation to address the challenge above. 
Instead of directly aligning the source and target domains, 
the proposed PrDA method introduces 
a sequence of intermediate virtual domains between the source and target domains in a progressive
manner through data interpolation,
and performs domain adaptation between the source domain and the virtual domain 
using multiple subspace alignment. 
It gradually augments the source domain with pseudo-labeled data from the virtual intermediate domains
after domain alignment 
and progressively moves the subspaces of source domain closer to the target domain. 
The alignment of the augmented source domain and the target domain is only 
conducted after the two domains become relatively much closer. 
Such a method is expected to be able to handle domain adaptation with large inter-domain variations.
We conduct experiments on multiple domain adaptation tasks and the experimental results
show the proposed approach achieves the state-of-the-art performance.

\section{Related Work}

In this section, we provide a brief review over unsupervised domain adaptation methods
and data interpolation based augmentation techniques. 

\subsection{Unsupervised Domain Adaptation}
Unsupervised domain adaptation addresses the setting where there are plenty of labeled instances
in the source domain while the data in the target domain is entirely unlabeled. 
Many unsupervised domain adaptation methods have been developed in the literature, including
divergence-based methods, adversarial-based methods, 
and subspace-based methods. 

Divergence oriented domain adaptation techniques focus on minimizing some criteria of domain divergence or discrepancy, 
such as Maximum Mean Discrepancy (MMD)~\cite{Long2015LearningNetworks}, Correlation Alignment (CORAL)~\cite{Sun2015ReturnAdaptation}, Contrastive Domain Discrepancy (CDD)~\cite{Kang2019ContrastiveAdaptation}, or Wasserstein distance~\cite{Damodaran2018DeepJDOT:Adaptation}, 
between the data distributions of the source and target domains to learn domain-invariant feature representations. 
The ultimate goal is such that the classifier trained on the source domain with the learned feature representation
can perform well on the target domain. 
Some work learns feature extractors by directly deploying multiple classifiers 
through Maximum Classifer Discrepancy (MCD)~\cite{Saito2018MaximumAdaptation}.
Avoiding expensive measures such as MMD, Central Moment Discrepancy (CMD)\cite{Zellinger2017CentralLearning} 
instead aims to match higher order moments of each domain. 

Adversarial-based methods extend the adversarial minimax training principle of 
Generative Adversarial Networks (GANs) \cite{Goodfellow2014GenerativeNets}
to align data distributions across domains. 
Such methods can include 
the generation of synthetic target data. 
The Coupled GAN (CoGAN) \cite{Liu2016CoupledNetworks} uses a pair of generators and discriminators for each domain, 
with some weight sharing between domains to learn a domain-invariant feature space. 
Some other approaches like the Domain Adversarial Neural Network (DANN) \cite{Ganin2017Domain-adversarialNetworksc} 
or the Reverse Gradient \cite{Ganin2014UnsupervisedBackpropagation} approach 
employ the use of a domain discriminator to learn domain invariant features. 
The Cycle-Consistent Adversarial Domain Adaptation (CyCADA) \cite{Hoffman2017CyCADA:Adaptationb} 
adapts representations at both the pixel and feature levels 
by transforming data across domains and conducting adversarial training.

Subspace-based techniques have also been used for domain adaptation
\cite{Fernando2014SubspaceAdaptation,Thopalli2019MultipleAdaptationb}, 
aiming to learn domain-invariant feature representations in a lower dimensional space. 
Domain adaptation is conducted then by aligning within the subspace through methods such as 
Geodesic Flow Kernel (GFK) \cite{Gong2012GeodesicAdaptation}, GFS \cite{Gopalan2011DomainApproach}, or by a learned transformation matrix \cite{Fernando2014SubspaceAdaptation}. Thopalli et al.~\cite{Thopalli2019MultipleAdaptationb} extend subspace-based domain adaptation by incorporating the concept of aligning multiple subspaces per domain with the advantage of each subspace being smaller in dimensions.

\subsection{Data Augmentation}
Given the omnipresent issue of lack of data, there has been much work on the field of data augmentation. 
Mixup is a fairly recent work that makes use of a very simple technique to augment the dataset with new 
interpolation samples \cite{Zhang2017Mixup:Minimization}. 
By generating linear interpolation points, Mixup can be used to generate virtual samples. 
MixMatch extends Mixup to leverage unlabeled data in a semi-supervised learning scenario \cite{Berthelot2019MixMatch:Learning}. 
It takes advantage of predicted labels as estimated pseudo-labels for the unlabeled portion of the dataset, 
then reweighs the losses to account for the estimation uncertainty. 
Such an approach to combining pseudo-labeling with standard Mixup opens the doors to a lot more interesting implementations of the basic premises that Mixup itself provides.

Inspired by Mixup and MixMatch, our proposed PrDA approach generates intermediate virtual domains between the source and target domains 
using the Mixup interpolation. 
It extends the multiple subspace technique in the context of unsupervised domain adaptation 
to help align domain pairs and 
generate additional data samples to augment our source dataset. 
To that end, our goal is to generalize the source domain distribution to overlap with the target domain distribution such that a classifier trained on the augmented source domain could learn domain invariant features. 
To ensure that a sufficient distribution of samples bridging the domain shift exist, we conduct the augmentation progressively: generated samples should start off strongly aligned towards the source domain distribution and gradually move towards the target domain distribution.

\begin{figure}[t]
    \centering
    \includegraphics[width=4in,height=2.1in]{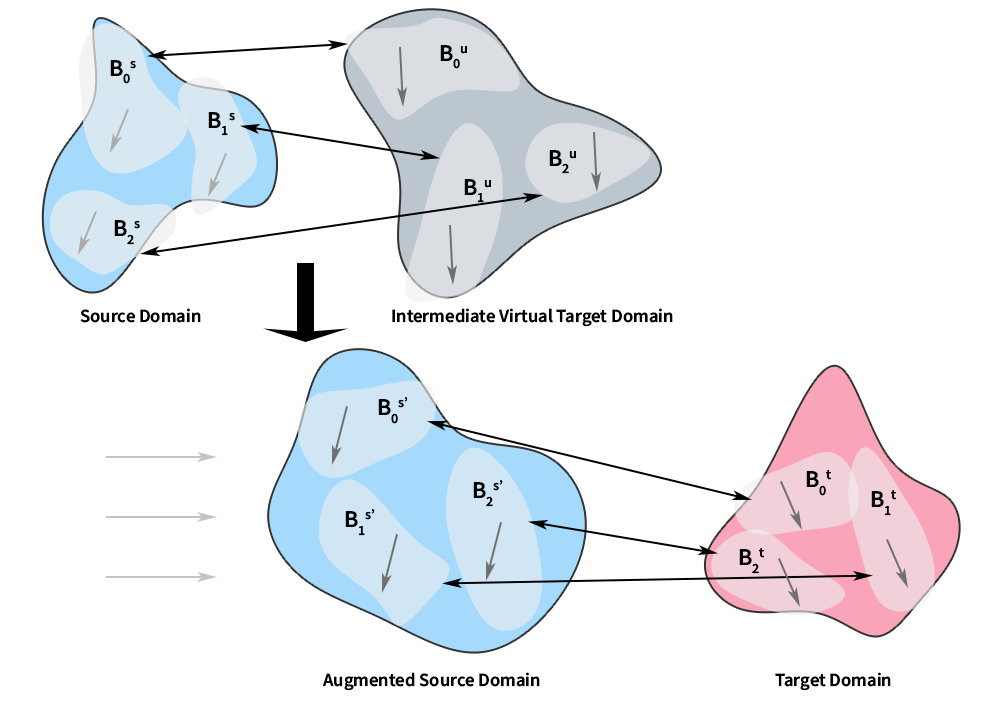}
    \caption{
Illustration of the progressive domain augmentation mechanism. 
It first generates a virtual domain between the source and target domains 
as a virtual target domain, 
and then aligns the subspaces of the source domain with the subspaces of the virtual target domain.
After alignment, the pseudo labels of the virtual samples can be predicted by the source classifier,
and the pseudo-labeled virtual samples can be selectively incorporated into the source domain
to form an augmented source domain, which now has reduced divergence from the original target domain.
    }
    \label{fig:model}
\end{figure}

\section{Method}

We consider the unsupervised domain adaptation setting, where we have a labeled source domain
and an unlabeled target domain.
Let $(X^s, Y^s)$ denote the labeled data in the source domain, 
where $X^s\in\mathbb{R}^{n_s\times d}$ is the sample feature matrix and  
$Y^s\in\{0,1\}^{n_s\times K}$ is the corresponding label indication matrix. 
Let $X^t\in\mathbb{R}^{n_t\times d}$ denote the unlabeled target domain data.
We assume there is substantial distribution divergence between the two domains,
and aim to bridge the domain divergence 
and learn a good classifier for the target domain. 

Instead of directly aligning the source and target domains within one step, 
we propose a novel progressive domain augmentation method
to gradually shift the source domain towards the target domain 
with data augmentation in multiple steps,
aiming to handle the domain divergence in a more subtle manner
by separating a large domain gap into multiple small ones.
The idea is that we use the interpolation point generation
method of Mixup to generate an auxiliary virtual domain between the source domain and target domain.
By using the generated virtual domain as an intermediate target domain, 
we can perform multiple subspace alignment to align the source domain with the virtual domain
and generalize the source trained classifier into the virtual domain to make predictions.
Virtual samples with confidently predicted pseudo-labels can then be used to augment the source domain,
which functions as a shifting of the source domain towards the target domain.
Figure~\ref{fig:model} illustrates this virtual domain assisted domain adaptation mechanism. 
This process can be repeated to generate a sequence of 
intermediate virtual domains and form a gradual shifting of the source domain. 
Moreover, we can also control the Mixup mechanism to make the generated virtual samples staying closer towards
the source domain in the early stage, while gradually moving closer to the target domain
in the later stage.
Below we present the virtual domain generation process, 
the multiple subspace alignment mechanism, and the overall algorithm.

\subsection{Virtual Domain Generation}
We adopt the Mixup linear data interpolation mechanism in~\cite{Zhang2017Mixup:Minimization} to generate virtual samples between two domains.
Mixup is defined as a form of generic vicinal distribution based on the theory of convex combination 
that aims to minimize a vicinal risk and construct a dataset of virtual samples. 
Given a set of samples $X$, for a randomly selected two samples $x_1,x_2\in X$
and their labels $y_1, y_2$, 
Mixup generates a new virtual sample $\hat{x}$ and its label $\hat{y}$ as follows:
\begin{align}
\hat{x} = \lambda x_1 + (1-\lambda)x_2,\qquad
\hat{y} = \lambda y_1 + (1-\lambda)y_2
\end{align}
where the parameter $\lambda$'s value is fairly important - 
although Zhang et al. propose to sample $\lambda$ from a $Beta$ distribution \cite{Zhang2017Mixup:Minimization}, 
the actual ramifications of the value of $\lambda$ are clear: higher $\lambda$ values favor $x_1$ and lower ones favor $x_2$. While this fact may be inconsequential for within-domain usage, it becomes an important consideration when addressing multiple domains and the problem of domain adaptation. 
With a slight modification, we extend the Eq.(1) to apply to the task of domain adaptation.
By sampling a source sample $x^s_i\in X^s$ 
and a target sample $x^t_j\in X^t$,  
we generate a virtual sample $\hat{x}$ inbetween:
\begin{align}
    \hat{x} = \lambda x^s_i + (1-\lambda)x^t_j
\end{align}
By generating a set of virtual samples with different source-target sample pairs,
we can form a virtual domain.  
As it will be used as an intermediate unlabeled target domain,
we do not require the labels to be generated.  
The importance of $\lambda$ becomes more apparent in this setting - a higher $\lambda$ 
generates a virtual domain closer to the source domain and 
a lower $\lambda$ favors the target domain. 
We exploit this fact to achieve a progressive move of the generated virtual samples 
from the source domain towards the target domain
by gradually decreasing the $\lambda$ value from an initial $\lambda_0>0.5$ 
in multiple steps.

\subsection{Multiple Subspace Generation and Alignment}
We propose to adopt a multiple subspace alignment technique~\cite{Thopalli2019MultipleAdaptationb}
to align the source domain with the virtual target domain.
The low dimensional subspaces of each domain can capture
the fundamental information and intrinsic geometry of the domain distribution.
Aligning the subspaces across domains will naturally 
lead to a geometric domain alignment,
while multiple subspace alignment on the Grassmann manifold can be more suitable 
for domain alignments than using a single set of subspaces 
\cite{Thopalli2019MultipleAdaptationb}.

The Grassmann Manifold $\mathbb{G}^{k,n}$ can be described as the collection of all possible $k$-dimensional 
linear subspaces in $\mathbb{R}^n$ \cite{Thopalli2019MultipleAdaptationb,Zhang2018GrassmannianLearning}. 
A special property of the manifold is that each point on the manifold can be represented as a basis set (i.e. a collection of orthonormal vectors). 
We assume
all subspaces for each domain must be $k$-dimensional and linear, and must respect the geometry of the Grassmann Manifold. 
Hence we can align subspaces across domains 
by exploiting the natural geometry of subspaces on the Grassmann manifold.

Given a dataset $X$, we use PCA to generate multiple sets of subspaces on the Grassmann Manifold.
First we perform PCA on $X$ to obtain its top-k principle components, $W\in\mathbb{R}^{d\times k}$,
such that $W^\top W = I$. 
$W$ forms 
a basis set $\mathcal{B}=\{b_1,b_2,..,b_k\}$, where each $b_i$ is a column of $W$.
This would be our solution for single set subspace generation. 
For multiple subspace generation, we need to continue generating the next set of subspaces
by focusing on data that were not explained well with the previous set of subspaces,
following~\cite{Thopalli2019MultipleAdaptationb}.
That is, we focus on the data samples that have high reconstruction errors with the
current basis set, $\|X_i- X_iWW^\top\|^2$.
Any sample with a high error (measured against a defined threshold $\tau$) 
can be allocated to a future subspace. We denote the high-error samples for $X$ as $X_e$.
The next subspace will be generated from $X_e$. 
Moreover, the current basis set (subspaces) can be refit to leave out $X_e$.
The overall multiple subspace generation process is shown in Algorithm~\ref{alg:msubspace}.
It induces a collection $\mathcal{M}$ of subspaces. 
\begin{algorithm}[t]
\caption{Multiple Subspace Generation}
	\label{alg:msubspace}
\begin{algorithmic}[1]
\Require {$X$: dataset, $k$, $\tau$: error threshold}
\State $\mathcal{M} :=$ initialize an empty collection to store subspaces for given dataset $X$
\Loop { until size of $X < k$}
\State $\mathcal{B} :=$ perform PCA on $X$, retaining the top $k$ principal components 
\State $X_e :=$ get the high-error samples with threshold $\tau$ on $X$ using $\mathcal{B}$
\State $\mathcal{B} :=$ refit subspace $\mathcal{B}$ using PCA on $X / X_e$ to get higher-fidelity subspace
\State $\mathcal{M} := \mathcal{M} \cup \mathcal{B}$ - add the subspace to the collection of subspaces $\mathcal{M}$
\State $X :=$ high-error samples $X_e$
\EndLoop
\end{algorithmic}
\end{algorithm}
We can apply the multiple subspace generation algorithm to both the source and virtual target domains, 
resulting in collections $\mathcal{M}^s$ and $\mathcal{M}^u$. 
This provides the foundation for cross domain subspace alignment.

\paragraph{\bf Aligning the Subspaces.}
The multiple subspace alignment task between 
$\mathcal{M}^s$ and $\mathcal{M}^u$ lies in matching a given source domain subspace $\mathcal{B}^s_i\in\mathcal{M}^s$ with a corresponding virtual domain subspace $\mathcal{B}^u_j\in\mathcal{M}^u$. 
This matching process relies on the subspaces being representable as points on the same Grassmann Manifold $\mathbb{G}^{k,n}$; we recall that each subspace is a basis set or orthonormal vectors. 
The main idea is that the closer points on the Grassmann manifold are easier to align. 
As such, we aim to measure the distance between the points. To accomplish this, we adopt a more general form of the symmetrical directional distance $d_\Delta$ \cite{Sun2007FurtherDistance}. For each basis vector $b^s_i\in\mathcal{B}^s$ and $b^u_j\in\mathcal{B}^u$ 
from a given subspace in each domain, we define the distance, also known as the chordal metric, as:
\begin{align}
    d_\Delta(\mathcal{B}^s,\mathcal{B}^u) = \Big(\max(r^s, r^u) - \sum_{i,j=1}^{r^s,r^u} (b^s_i\text{}^\intercal\cdot b^u_j)^2\Big)^{\frac{1}{2}}
\end{align}
where $r^s$ and $r^u$ represent the rank of the basis sets $\mathcal{B}^s$ and $\mathcal{B}^u$ respectively. Because each basis set is orthonormal, we can consider each subspace to be an orthogonal matrix. As such, each subspace would have full rank (i.e. rank($\mathcal{B}^s$)$ = k$). Based on our design, both source and (virtual) target domain subspaces have the same $k$ value, so we can assume that $r^s=r^u$. With this in mind, we can further reduce our distance formula to:
\begin{align}
    d_\Delta(\mathcal{B}^s,\mathcal{B}^u) = \Big(k - \sum_{i,j=1}^{k,k} (b^s_i\text{}^\intercal\cdot b^u_j)^2\Big)^{\frac{1}{2}}
	\label{chordal}
\end{align}

We first match the two subspaces with the smallest distance, then remove them and find the next two.
Once each subspace has a corresponding matching subspace, a final transformation is required to align the source domain subspace with its corresponding virtual target domain subspace. This transformation turns out to be extremely simply by virtue of being on the Grassmann Manifold - we can align them through a linear transformation. We define a transformation matrix $A$. 
For each $\mathcal{B}^s_i\in\mathcal{M}^s$ and $\mathcal{B}^u_j\in\mathcal{M}^u$, we aim to minimize the following objective \cite{Fernando2014SubspaceAdaptation}:
\begin{align}
    A^* &= \text{arg}\min\limits_A||\mathcal{B}^s_i A - \mathcal{B}^u_j||_F
    = (\mathcal{B}^s_i)^\intercal\mathcal{B}^u_j
\end{align}
where $||.||_F$ denotes the Frobenius norm and $\mathcal{B}^s_i$ and $\mathcal{B}^u_j$ are matched subspaces. Given this transformation matrix $A$, we can thus transform the source domain $\mathcal{B}^s_i$ to a corresponding virtual target domain $\mathcal{B}^u_j$ by multiplying the subspace with the calculated transformation matrix:
\begin{align}
    \mathcal{B}^{ta}_i = \mathcal{B}^s_i \cdot A^*
\end{align}
This new resultant subspace, $\mathcal{B}^{ta}_i$, is known as the 'target aligned' subspace. This process is repeated for all pairs of matching subspaces calculated in the previous step. Using these, we can align our source domain dataset $X^s$ to the virtual target domain by projecting each with their associated newly calculated 'target aligned' subspace:
\begin{align}
    Z^s = X^s \cdot \mathcal{B}^{ta}_i
\end{align}
The alignment process for the virtual target domain dataset is far simpler - because we aren't shifting anything, we can simply project the 
data samples to their associated subspaces:
\begin{align}
    Z^u = X^u \cdot \mathcal{B}^u_j
\end{align}

\subsection{Overall Training Algorithm}

To summarize, the proposed approach generates a virtual domain between the source and target domains
by deploying Mixup sample generation. 
The virtual domain is then used as an intermediate target domain to be aligned with the source domain
through multiple subspace alignment. 
After alignment, the data in both domains can be projected into the aligned subspaces, 
where the trained source domain classifier over the aligned subspaces can be used to predict the labels for the instances
in the virtual target domain.
From these virtual domain samples and their predicted pseudo-labels, we select those samples who's pseudo-labels are above a user-defined threshold $\rho$ and use them to augment the source domain data, 
This process can be repeated to generate multiple virtual domains to progressively shift the source domain towards the target domain,
while the $\lambda$ is also used to control to shifting phase.
The overall training algorithm is presented in Algorithm~\ref{alg:train}.

\begin{algorithm}[t]
\caption{Training Algorithm}
	\label{alg:train}
\begin{algorithmic}[1]
\Require {$X^s, Y^s$: labelled source dataset, $X^t$: unlabeled target dataset, 
\Require $\lambda_\text{set}$: $[0.8,0.6, 0.4, 0.2]$}
\State $F :=$ initialize a classifier
\State $G :=$ initialize a feature extractor
\State Pretrain classifier $F$ and feature extractor $G$ on $(X^s, Y^s)$ with cross-entropy loss
	\State $\mathcal{M}^s :=$ generate collection of subspaces for $G(X^s)$ for a given $k$ and $\tau$ threshold
\For {$\lambda \in \lambda_\text{set}$}
\State $X^u:=$ initialize an empty set to store new virtual samples
\State $X^s_\Delta:=$ randomly sample a batch from $X^s$
\State $X^t_\Delta:=$ randomly sample a batch from $X^t$ 
\For {$i,j\in\Delta$}
\State $X^u:=X^u\cup (\lambda x^s_i + (1-\lambda)x^t_j)$ - generate a virtual sample 
\EndFor
	\State $\mathcal{M}^u :=$ generate collection of subspaces for $G(X^u)$ for a given $k$ and $\tau$ threshold
\State Match each subspace in $\mathcal{M}^s$ with a subspace in $\mathcal{M}^{u}$ 
	based on distance in Eq.(\ref{chordal})
\State $\mathcal{M}^{ta} :=$ compute target-aligned subspace for all the subspaces in $\mathcal{M}^s$ 
\State $Z^s :=$ project source data $X^s$ to target-aligned subspaces $\mathcal{M}^{ta}$
\State $Z^u :=$ project target data $X^u$ to target subspaces in $\mathcal{M}^{u}$
\State Train classifier $H$ on projected source domain $(Z^s, Y^s)$
\State $\hat{Y}^u:=$ predict labels for virtual dataset $X^u$ using classifier $H$
\State $(X^s, Y^s):=(X^s, Y^s)\cup(X^u, \hat{Y}^u)$ - augment source domain data with selected virtual samples and their pseudo-labels s.t. pseudo-label confidence is $>\rho$
\State Update $\mathcal{M}^s$ based on updated $X^s$ 
\State Train classifier $F$ and $G$ on augmented $(X^s, Y^s)$ using cross-entropy loss
\EndFor
\end{algorithmic}
\end{algorithm}


\section{Experiments}
We evaluate our proposed PrDA approach on three standard unsupervised domain adaptation tasks: \textit{Office-31} \cite{Saenko2010AdaptingDomains}, \textit{Office-Caltech-10} \cite{Gong2012GeodesicAdaptation}, and \textit{ImageCLEF-DA}\footnote{http://imageclef.org/2014/adaptation}. We compare the results of our approach to a number of recent state-of-the-art methods, focusing, when possible, on similar
transformation-based
learning models. 
In this section, we report the experimental setting and results.

\subsection{Datasets}
\begin{figure}[t]
    \centering
    \includegraphics[width=4in,height=2.5in]{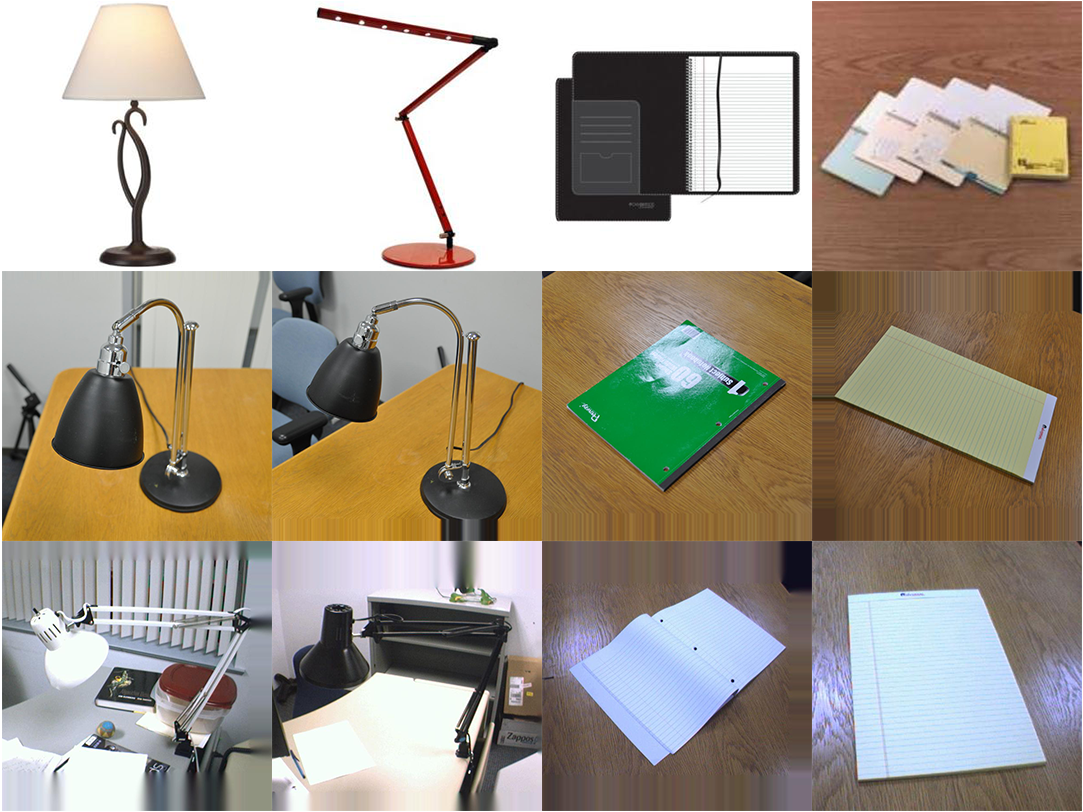}
    \caption{\textbf{Office-31 Dataset} - Four samples from each of the three 
	domains. 
	Starting from the top downwards: (1) Amazon (2) DLSR (3) Webcam.}
    \label{fig:office31-sample}
\end{figure}

\paragraph{\bf Office-31.}
This dataset is an extremely popular dataset for visual domain adaptation. It features 31 different classes, for a total of 4,652 images, and includes three different domains: Amazon (\textbf{A}), DSLR (\textbf{D}), and Webcam (\textbf{W}). 
Figure~\ref{fig:office31-sample} shows some sample images in these three domains.
As is standard, we evaluate our method 
on all the domain adaptation tasks constructed with permutations of these domains: $\textbf{A}\rightarrow \textbf{W}$, $\textbf{W}\rightarrow \textbf{A}$, $\textbf{A}\rightarrow \textbf{D}$, $\textbf{D}\rightarrow \textbf{A}$, $\textbf{D}\rightarrow \textbf{W}$, and $\textbf{W}\rightarrow \textbf{D}$.

\paragraph{\bf Office-Caltech-10.}
This dataset is a combination of the Office-31 dataset and the Caltech-256 dataset, whereby the 10 overlapping classes are used between them. 
With the inclusion of Caltech-256, this dataset includes four different domains: Amazon (\textbf{A}), DSLR (\textbf{D}), Webcam (\textbf{W}), and Caltech (\textbf{C}). 
This results in 2,505 images across the four domains.
As with the Office-31 dataset, we evaluate on all 12 permutations of these domains.

\paragraph{\bf ImageCLEF-DA.}
The dataset is a specialized subset consisting of 12 overlapping classes from three public datasets: Caltech-256 (\textbf{C}), ImageNet ILSVRC2012 (\textbf{I}), and Pascal VOC 2012 (\textbf{P}). 
As per standard protocol for the ImageCLEF-DA task, each class contains 50 random images from 
the category.
Same as with the two other datasets, we evaluate on all 6 permutations of these three domains.

\subsection{Setup}
We adopt a standard testing protocol by using ResNet-50 \cite{He2015DeepRecognition} as the back-bone feature extractor for all tasks, fine-tuned on the source domain dataset. Our experiments follow standard unsupervised domain adaptation protocol: all labeled source samples and all unlabeled target samples are used. Reported classification accuracy for each task is an average of 10 repeated runs. For all experiments, we use the same set of hyperparameters: subspace dimension $k=44$, subspace error threshold $\tau$ is set to be within the range $[0.1,0.4]$ (dependent on the particular task), and the $\lambda$ value controlling which domain is favored is initially set at $0.8$ and progressively decreased over time by a unit of 0.2. 
As in \cite{Thopalli2019MultipleAdaptationb}, we found the typical number of subspaces per domain tended to be around 3 or 4.

\subsection{Experiment Results}
\subsubsection{Office-31}
The results on the \textit{Office-31} set of tasks are reported in Table \ref{tab:office31}. We note that our results far exceed those of the base subspace alignment technique SA in every task, showing the strength of our proposed PrDA approach. 
We compare to other state-of-the-art transformation or subspace based learning methods, including the CORrelation ALignment (CORAL) \cite{Sun2016CorrelationAdaptation}, which uses a transformation matrix to minimize covariance distance between domains, RWA \cite{vanLaarhoven2018UnsupervisedLabelings}, which uses a random walk based algorithm,
and MEDA \cite{Wang2018VisualAlignment}, which is an approach centered on learning a domain-invariant classifier in the Grassmann Manifold. 
Our results surpass all compared techniques in the $A\rightarrow D$ task. We also note that our results for the $W\rightarrow D$ task is very close to matching the reported state-of-the-art. 

\begin{table}[t]
    \centering
	\setlength\tabcolsep{5pt}
    \caption{\textbf{Office-31} results for each of the 6 transfer tasks in the \textit{Office-31} dataset. Results all use \textit{ResNet-50} as the feature extractor, and all results are averages.}
    \begin{tabular}{l|c|c|c|c|c|c}
        \hline Method & $A\rightarrow W$ & $W\rightarrow A$ & $A\rightarrow D$ & $D\rightarrow A$ & $D\rightarrow W$ & $W\rightarrow D$\\\hline
        SA\cite{Fernando2014SubspaceAdaptation} & 76.7 & 61.8 & 75.5 & 59.6 & 93.6 & 95.8\\
        CORAL\cite{Sun2016CorrelationAdaptation} & 77.6 & 59.9 & 80.9 & 58.8 & 98.6 & \textbf{100.0}\\
        RWA\cite{vanLaarhoven2018UnsupervisedLabelings} & \textbf{90.6} & 73.7 & 90.0 & \textbf{74.4} & \textbf{99.0} & 99.6\\
        MEDA\cite{Wang2018VisualAlignment} & 86.2 & \textbf{74.0} & 85.3 & 72.4 & 97.2 & 99.4\\\hline
        PrDA & 88.1 & 68.9 & \textbf{92.2} & 71.5 & 97.1 & 99.6\\\hline
    \end{tabular}
    \label{tab:office31}
\end{table}

To measure if our proposed PrDA approach excels with larger cross-domain divergence, 
we measured the similarity of each pair of domains in each task. We set up a basic two-class classification scenario, with each domain's samples categorized as a different class. We shuffle these samples together and perform 5-fold cross-validation to get an accurate estimation of the model's accuracy overall. We interpret a high accuracy result for a given task as indicative 
of a larger dissimilarity between domains (larger domain divergence), as the classifier was able to more accurately separate the two distributions. Inversely, we consider a lower accuracy result as indicative of a greater degree of similarity between the domains, as the classifier is unable to confidently separate the two distributions. Table \ref{tab:office31abla} summarizes the average classification accuracy obtained for each task. We note that the higher accuracy values obtained for a task in this ablation study match tasks in which we obtained state-of-the-art or closer to state-of-the-art results. These results bolster our hypothesis 
that our proposed PrDA approach's technique of separating the alignment process into several smaller ones achieves superior results for greater divergences between domains.
\begin{table}[t]
    \centering
	\setlength\tabcolsep{5pt}
    \caption{Domain classification results on the tasks of \textbf{Office-31}
	Results are taken from a domain classifier attempting to classify a sample as part of one domain or the other, and results are based on 5-fold cross-validation.
	Larger values indicate larger domain divergence. }
    \begin{tabular}{l|c|c|c|c|c|c}
         \hline Task & $A\rightarrow W$ & $W\rightarrow A$ & $A\rightarrow D$ & $D\rightarrow A$ & $D\rightarrow W$ & $W\rightarrow D$\\\hline
         Accuracy & 52.6 & 52.6 & 75.5 & 75.5 & 55.0 & 55.0\\\hline
    \end{tabular}
    \label{tab:office31abla}
\end{table}

\subsubsection{Office-Caltech-10}
The results for the \textit{Office-Caltech-10} set of tasks are reported in Table \ref{tab:office-caltech}. Once again, we note significant improvements as compared to the base subspace alignment technique SA, further enforcing the strengths of our proposed PrDA approach to domain alignment. As compared to Ad-REM \cite{vanLaarhoven2018DomainMaximization}, a technique centered on randomized expectation maximization, our approach achieves state-of-the-art results in several of the tasks. We further note that our results for tasks that were not state-of-the-art were still very competitive. 
Overall, our proposed PrDA produces the best results on 7 out of the total 12 tasks among
all the comparison methods
and demonstrates effective performance.

\begin{table}[t]
    \centering
	\setlength\tabcolsep{1pt}
    \caption{\textbf{Office-Caltech-10} results for each of the 12 transfer tasks in the \textit{Office-Caltech-10 dataset}. Results all use \textit{ResNet-50} as the feature extractor, and all reported results are averages over ten runs.}
    \begin{tabular}{l|c|c|c|c|c|c|c|c|c|c|c|c}
         \hline Method & $A\cdot C$ & $A\cdot D$ & $A\cdot W$ & $C\cdot A$ & $C\cdot D$ & $C\cdot W$ & $D\cdot A$ & $D\cdot C$ & $D\cdot W$ & $W\cdot A$ & $W\cdot C$ & $W\cdot D$\\\hline
         SA\cite{Fernando2014SubspaceAdaptation} & 88.6 & 89.9 & 88.3 & 93.2 & 90.4 & 88.8 & 88.4 & 85.0 & 97.8 & 89.5 & 84.3 & 99.9\\
         CORAL\cite{Sun2016CorrelationAdaptation} & 88.8 & 93.2 & 90.5 & 94.2 & 92.5 & 90.8 & 92.9 & 87.4 & 98.1 & 92.1 & 85.9 & \textbf{100.0}\\
         Ad-REM\cite{vanLaarhoven2018DomainMaximization} & 93.4 & 98.9 & 98.9 & 95.6 & 98.1 & 97.2 & \textbf{96.0} & \textbf{94.3} & 98.6 & \textbf{95.9} & \textbf{93.7} & \textbf{100.0}\\
         RWA\cite{vanLaarhoven2018UnsupervisedLabelings} & \textbf{93.8} & 98.9 & 97.8 & 95.3 & \textbf{99.4} & 95.9 & 95.8 & 93.1 & 98.4 & 95.3 & 92.4 & 99.9\\\hline
         PrDA & 92.1 & \textbf{99.0} & \textbf{99.3} & \textbf{97.2} & \textbf{99.4} & \textbf{98.3} & 94.7 & 91.0 & \textbf{99.7} & 95.6 & 93.4 & \textbf{100.0}\\\hline 
    \end{tabular}
    \label{tab:office-caltech}
\end{table}

Once again, to verify our claim with respect to our performance when considering high cross-domain divergence, we conduct a study to measure the similarity of each pair of domains in each task. We set up the same two-class classification scenario as with the \textit{Office-31} set of tasks, with each domain in the pair a different class. Given the same interpretation as with \textit{Office-31}, we note overall higher reported accuracy values, indicating that, in general, a larger degree of domain divergence exists per pair of domains. These results corroborate our approach's performance in this set of tasks, further emphasizing our approach's capacity in handling
large domain divergence gaps.

\begin{table}[htb]
    \centering
    \caption{Domain classification accuracy on \textbf{Office-Caltech-10}
	Results are taken from a domain classifier attempting to classify a sample as part of one domain or the other, and results are based on 5-fold cross-validation.
	Larger accuracy values indicate larger domain divergence.
	}
    \begin{tabular}{l|c|c|c|c|c|c|c|c|c|c|c|c}
         \hline Task & $A\cdot C$ & $A\cdot D$ & $A\cdot W$ & $C\cdot A$ & $C\cdot D$ & $C\cdot W$ & $D\cdot A$ & $D\cdot C$ & $D\cdot W$ & $W\cdot A$ & $W\cdot C$ & $W\cdot D$\\\hline
  Accuracy & 50.6 & 76.1 & 62.9 & 50.6 & 76.1 & 62.9 & 76.1 & 76.1 & 65.3 & 62.9 & 62.9 & 65.3\\\hline
    \end{tabular}  
\label{tab:office-caltech-abla}
\end{table}

\subsubsection{ImageCLEF-DA}
The results for the \textit{ImageCLEF-DA} set of tasks are reported in Table \ref{tab:imageclef}. We compare our approach's performance to 
MEDA \cite{Wang2018VisualAlignment}, an approach centered on learning a domain-invariant classifier in the Grassmann Manifold, 
Conditional Domain Adversarial Network (CDAN) \cite{Long2018ConditionalAdaptation}, an adversarial approach with additional conditioning, and Domain Adversarial Neural Network (DANN) \cite{Ganin2016Domain-AdversarialNetworks}, which leverages domain discriminant and domain invariant features in an adversarial setting. 
The tasks on this dataset have small domain divergence values according to our two-class domain separation study.
Nevertheless, the proposed PrDA produces the best results on 3 of the 6 transfer tasks,
and achieves comparable performance on 2 other tasks.

\begin{table}[t]
    \centering
	\setlength\tabcolsep{5pt}
    \caption{\textbf{ImageCLEF-DA} results for each of the 6 transfer tasks in the \textit{ImageCLEF-DA} dataset. Results all use \textit{ResNet-50} as the feature extractor, and all results are averages over ten runs.}
    \begin{tabular}{l|c|c|c|c|c|c}
        \hline Method & $I\rightarrow P$ & $P\rightarrow I$ & $I\rightarrow C$ & $C\rightarrow I$ & $C\rightarrow P$ & $P\rightarrow C$\\\hline
        DANN\cite{Ganin2016Domain-AdversarialNetworks} & 75.0 & 86.0 & 96.2 & 87.0 & 74.3 & 91.5\\
        CDAN\cite{Long2018ConditionalAdaptation} & 76.7 & 90.6 & 97.0 & 90.5 & 74.5 & 93.5\\
        CDAN+E\cite{Long2018ConditionalAdaptation} & 77.7 & 90.7 & \textbf{97.7} & 91.3 & 74.2 & 94.3\\
        MEDA\cite{Wang2018VisualAlignment} & 79.7 & \textbf{92.5} & 95.7 & \textbf{92.2} & 78.5 & \textbf{95.5}\\\hline
        PrDA & \textbf{81.8} & 91.0 & 94.5 & 88.2 & \textbf{80.5} & \textbf{95.5}\\\hline
    \end{tabular}
    \label{tab:imageclef}
\end{table}

\section{Conclusion}
This paper presented a novel Progressive Data Augmentation (PrDA) approach for unsupervised domain adaptation 
based on multiple subspace alignment.
It deploys Mixup data interpolation to generate a sequence of intermediate virtual domains to 
progressively align the source domain with the target domain, 
resulting in an approach that effectively divides a larger domain shift into a series of smaller alignment problems. This is in contrast with 
the majority existing works that tends 
to perform the full alignment at once. 
We demonstrate that our approach is particularly viable for larger domain shifts via a combination of experiments and additional studies. While simple in design, our approach demonstrates state-of-the-art performance on several standard domain adaptation tasks, validating our progressive approach towards domain adaptation.

\bibliographystyle{splncs04.bst}
\bibliography{references.bib}

\begin{thebibliography}{10}
\providecommand{\url}[1]{\texttt{#1}}
\providecommand{\urlprefix}{URL }
\providecommand{\doi}[1]{https://doi.org/#1}

\bibitem{Baktashmotlagh2013UnsupervisedProjection}
Baktashmotlagh, M., Harandi, M.T., Lovell, B.C., Salzmann, M.: {Unsupervised
  domain adaptation by domain invariant projection}. In: Proceedings of the
  IEEE International Conference on Computer Vision. pp. 769--776 (2013)

\bibitem{Beijbom2012DomainApplications}
Beijbom, O.: {Domain Adaptations for Computer Vision Applications}. Tech. rep.
  (11 2012)

\bibitem{Berthelot2019MixMatch:Learning}
Berthelot, D., Carlini, N., Goodfellow, I., Papernot, N., Oliver, A., Raffel,
  C.: {MixMatch: A Holistic Approach to Semi-Supervised Learning}. Advances in
  Neural Information Processing Systems 32  (5 2019)

\bibitem{Damodaran2018DeepJDOT:Adaptation}
Damodaran, B.B., Kellenberger, B., Flamary, R., Tuia, D., Courty, N.:
  {DeepJDOT: Deep Joint Distribution Optimal Transport for Unsupervised Domain
  Adaptation}. European Conference on Computer Vision  (3 2018)

\bibitem{Fernando2014SubspaceAdaptation}
Fernando, B., Habrard, A., Sebban, M., Tuytelaars, T.: {Subspace Alignment For
  Domain Adaptation}. Tech. rep. (9 2014)

\bibitem{Ganin2014UnsupervisedBackpropagation}
Ganin, Y., Lempitsky, V.: {Unsupervised Domain Adaptation by Backpropagation}.
  International Conference on Machine Learning  (9 2014)

\bibitem{Ganin2016Domain-AdversarialNetworks}
Ganin, Y., Ustinova, E., Ajakan, H., Germain, P., Larochelle, H., Laviolette,
  F., Marchand, M., Lempitsky, V.: {Domain-Adversarial Training of Neural
  Networks}. Journal of Machine Learning  (5 2016)

\bibitem{Ganin2017Domain-adversarialNetworksc}
Ganin, Y., Ustinova, E., Ajakan, H., Germain, P., Larochelle, H., Laviolette,
  F., Marchand, M., Lempitsky, V.: {Domain-adversarial training of neural
  networks}. In: Advances in Computer Vision and Pattern Recognition, pp.
  189--209. Springer London (2017)

\bibitem{Ghifary2016DeepAdaptationb}
Ghifary, M., Kleijn, W.B., Zhang, M., Balduzzi, D., Li, W.: {Deep
  Reconstruction-Classification Networks for Unsupervised Domain Adaptation}.
  European Conference on Computer Vision  (7 2016)

\bibitem{Gong2012GeodesicAdaptation}
Gong, B., Shi, Y., Sha, F., Grauman, K.: {Geodesic Flow Kernel for Unsupervised
  Domain Adaptation}. Proceedings of the IEEE Conference on Computer Vision and
  Pattern Recognition  (2012)

\bibitem{Goodfellow2014GenerativeNets}
Goodfellow, I.J., Pouget-Abadie, J., Mirza, M., Xu, B., Warde-Farley, D.,
  Ozair, S., Courville, A., Bengio, Y.: {Generative Adversarial Nets}.
  Conference on Neural Information Processing Systems  (2014)

\bibitem{Gopalan2011DomainApproach}
Gopalan, R., Li, R., Chellappa, R.: {Domain Adaptation for Object Recognition:
  An Unsupervised Approach}. International Conference on Computer Vision
  (2011)

\bibitem{Gu2019ImprovingInformation}
Gu, S., Feng, Y., Liu, Q.: {Improving Domain Adaptation Translation with Domain
  Invariant and Specific Information}. Proceedings of NAACL-HLT pp. 3081--3091
  (2019)

\bibitem{He2015DeepRecognition}
He, K., Zhang, X., Ren, S., Sun, J.: {Deep Residual Learning for Image
  Recognition}. Tech. rep. (12 2015)

\bibitem{Hoffman2013EfficientRepresentations}
Hoffman, J., Rodner, E., Donahue, J., Darrell, T., Saenko, K.: {Efficient
  Learning of Domain-invariant Image Representations}. ICLR  (2013)

\bibitem{Hoffman2017CyCADA:Adaptationb}
Hoffman, J., Tzeng, E., Park, T., Zhu, J.Y., Isola, P., Saenko, K., Efros,
  A.A., Darrell, T.: {CyCADA: Cycle-Consistent Adversarial Domain Adaptation}.
  PMLR  (11 2017)

\bibitem{Kang2019ContrastiveAdaptation}
Kang, G., Jiang, L., Yang, Y., Hauptmann, A.G.: {Contrastive Adaptation Network
  for Unsupervised Domain Adaptation}. CVPR 2019  (2019)

\bibitem{Kumar2018Co-regularizedAdaptation}
Kumar, A., Sattigeri, P., Wadhawan, K., Karlinsky, L., Feris, R., Freeman,
  W.T., Wornell, G.: {Co-regularized Alignment for Unsupervised Domain
  Adaptation}. NeurIPS  (2018)

\bibitem{vanLaarhoven2018DomainMaximization}
van Laarhoven, T., Marchiori, E.: {Domain Adaptation with Randomized
  Expectation Maximization}  (3 2018)

\bibitem{vanLaarhoven2018UnsupervisedLabelings}
van Laarhoven, T., Marchiori, E.: {Unsupervised Domain Adaptation with Random
  Walks on Target Labelings}. Tech. rep. (6 2018)

\bibitem{Li2019Cycle-consistentNetworks}
Li, J., Zhu, L., Chen, E., Lu, K., Ding, Z., Huang, Z.: {Cycle-consistent
  conditional adversarial transfer networks}. In: MM 2019 - Proceedings of the
  27th ACM International Conference on Multimedia. pp. 747--755. Association
  for Computing Machinery, Inc (10 2019)

\bibitem{Liu2016CoupledNetworksb}
Liu, M.Y., Tuzel, O.: {Coupled Generative Adversarial Networks}. Neural
  Information Processing Systems (NIPS)  (6 2016)

\bibitem{Liu2016CoupledNetworks}
Liu, M.Y., Tuzel, O.: {Coupled Generative Adversarial Networks}  (6 2016)

\bibitem{Long2015LearningNetworks}
Long, M., Cao, Y., Wang, J., Jordan, M.I.: {Learning Transferable Features with
  Deep Adaptation Networks}  (2 2015)

\bibitem{Long2018ConditionalAdaptation}
Long, M., Cao, Z., Wang, J., Jordan, M.I.: {Conditional Adversarial Domain
  Adaptation}. NeurIPS  (5 2018)

\bibitem{Morerio2020GenerativeAdaptation}
Morerio, P., Volpi, R., Ragonesi, R., Murino, V.: {Generative Pseudo-label
  Refinement for Unsupervised Domain Adaptation}. Tech. rep. (1 2020)

\bibitem{Saenko2010AdaptingDomains}
Saenko, K., Kulis, B., Fritz, M., Darrell, T.: {Adapting Visual Category Models
  to New Domains}. European Conference on Computer Vision (ECCV)  (2010)

\bibitem{Saito2018MaximumAdaptation}
Saito, K., Watanabe, K., Ushiku, Y., Harada, T.: {Maximum Classifier
  Discrepancy for Unsupervised Domain Adaptation}  (12 2018)

\bibitem{Shimodaira2000ImprovingFunction}
Shimodaira, H.: {Improving predictive inference under covariate shift by
  weighting the log-likelihood function}. Journal of Statistical Planning and
  Inference  \textbf{90}(2),  227--244 (10 2000)

\bibitem{Sun2015ReturnAdaptation}
Sun, B., Feng, J., Saenko, K.: {Return of Frustratingly Easy Domain Adaptation}
   (11 2015)

\bibitem{Sun2016CorrelationAdaptation}
Sun, B., Feng, J., Saenko, K.: {Correlation Alignment for Unsupervised Domain
  Adaptation}  (12 2016)

\bibitem{Sun2007FurtherDistance}
Sun, X., Wang, L., Feng, J.: {Further Results on the Subspace Distance}.
  Pattern Recognition  (2007)

\bibitem{Thopalli2019MultipleAdaptationb}
Thopalli, K., Anirudh, R., Thiagarajan, J.J., Turaga, P.: {Multiple Subspace
  Alignment Improves Domain Adaptation}. ICASSP  (11 2019)

\bibitem{Tommasi2012LearningKnowledge}
Tommasi, T.: {Learning to Learn by Exploiting Prior Knowledge}. Ph.D. thesis,
  {\'{E}}cole Polytechnique F{\'{e}}d{\'{e}}rale de Lausanne (2012)

\bibitem{Wang2018VisualAlignment}
Wang, J., Feng, W., Chen, Y., Huang, M., Yu, H., Yu, P.S.: {Visual domain
  adaptation with manifold embedded distribution alignment}. In: MM 2018 -
  Proceedings of the 2018 ACM Multimedia Conference. pp. 402--410. Association
  for Computing Machinery, Inc (10 2018)

\bibitem{WangDeepSurvey}
Wang, M., Deng, W.: {Deep Visual Domain Adaptation: A Survey}. Tech. rep.

\bibitem{YuTransferNetwork}
Yu, C., Wang, J., Chen, Y., Huang, M., Key, B.: {Transfer Learning with Dynamic
  Adversarial Adaptation Network}. Tech. rep.

\bibitem{Zellinger2017CentralLearning}
Zellinger, W., Grubinger, T., Lughofer, E., Natschl{\"{a}}ger, T.,
  Saminger-Platz, S.: {Central Moment Discrepancy (CMD) for Domain-Invariant
  Representation Learning}. ICLR  (2 2017)

\bibitem{Zhang2017Mixup:Minimization}
Zhang, H., Cisse, M., Dauphin, Y.N., Lopez-Paz, D.: {mixup: Beyond Empirical
  Risk Minimization}  (10 2017)

\bibitem{Zhang2018GrassmannianLearning}
Zhang, J., Zhu, G., Heath, R.W., Huang, K.: {Grassmannian Learning: Embedding
  Geometry Awareness in Shallow and Deep Learning}  (8 2018)

\bibitem{Zhang2013DomainShift}
Zhang, K., Sch{\"{o}}lkopf, B., Wang, Z.: {Domain Adaptation under Target and
  Conditional Shift}. Internation Conference on Machine Learning  (2013)

\bibitem{Zhao2019OnAdaptation}
Zhao, H., Combes, R.T.d., Zhang, K., Gordon, G.J.: {On Learning Invariant
  Representation for Domain Adaptation}. International Conference on Machine
  Learning  (1 2019)

\bibitem{Zhu2017UnpairedNetworksc}
Zhu, J.Y., Park, T., Isola, P., Efros, A.A.: {Unpaired Image-to-Image
  Translation using Cycle-Consistent Adversarial Networks}. ICCV  (3 2017)

\end{thebibliography}
\end{document}